\documentclass[letterpaper, 10 pt, conference, nofonttune]{ieeeconf} 

\IEEEoverridecommandlockouts                              

\usepackage{amsthm}

\usepackage{subfigure}
\usepackage{xcolor}
\usepackage{times}
\usepackage{epsfig}
\usepackage{amsmath}
\usepackage{amssymb}
\usepackage{mathrsfs}
\usepackage[ruled,vlined,linesnumbered]{algorithm2e}
\usepackage{wrapfig}

\newcommand{\RRR}{\mathbb{R}}

\newcommand{\GGG}{\mathcal{G}}
\newcommand{\RU}{\mathcal{R}}
\newcommand{\NNN}{\mathcal{N}}

\newcommand{\EEE}{\mathcal{E}}
\newcommand{\XXX}{\mathcal{X}}
\newcommand{\ZZZ}{\mathcal{Z}}

\newcommand{\MMM}{\mathcal{M}}

\newcommand{\PPP}{\mathcal{P}}

\newcommand{\PP}{\mathbf{P}}

\newcommand{\QQ}{\mathbf{Q}}
\newcommand{\KK}{\mathbf{K}}
\newcommand{\II}{\mathbf{I}}
\newcommand{\RR}{\mathbf{R}}
\newcommand{\WW}{\mathbf{W}}
\newcommand{\EE}{\mathbf{E}}

\newcommand{\vv}{\mathbf{v}}
\newcommand{\xx}{\mathbf{x}}
\newcommand{\yy}{\mathbf{y}}
\newcommand{\zz}{\mathbf{z}}
\newcommand{\mm}{\mathbf{m}}

\newcommand{\st}{\mathbf{s}}

\newcommand{\ee}{\mathbf{e}}

\title{\LARGE \bf
Asynchronous Collaborative Localization by Integrating Spatiotemporal Graph Learning with Model-Based Estimation
}

\begin{document}
\author{Peng Gao$^{1}$, Brian Reily$^{1}$, Rui Guo$^{2}$, Hongsheng Lu$^{2}$, Qingzhao Zhu$^{1}$ and Hao Zhang$^{1}$
\thanks{$^{1}$Peng Gao, Brian Reily, Qingzhao Zhu and Hao Zhang are with the Human-Centered Robotics Laboratory at the Colorado School
of Mines, Golden, CO 80401, USA. $\{$gaopeng, hzhang$\}$@mines.edu}%
\thanks{$^{2}$Rui Guo and Hongsheng Lu are with Toyota Motor North America, Mountain View, CA 94043. $\{$rui.guo, hongsheng.lu$\}$@toyota.com}%
}

\maketitle
\thispagestyle{empty}
\pagestyle{empty}

\begin{abstract}
Collaborative localization is an essential capability for a team of robots such as connected vehicles to collaboratively estimate object locations from multiple perspectives with reliant cooperation.
To enable collaborative localization, four key challenges must be addressed,
including modeling complex relationships between observed objects,
fusing observations from an arbitrary number of collaborating robots,
quantifying localization uncertainty,
and addressing latency of robot communications.
In this paper, we introduce a novel approach
that integrates uncertainty-aware spatiotemporal graph learning and model-based state estimation
for a team of robots to collaboratively localize objects.
Specifically, we introduce a new uncertainty-aware graph learning model
that learns spatiotemporal graphs to represent historical motions of the objects observed by each robot over time and provides uncertainties in object localization.
Moreover, we propose a novel method for integrated learning and model-based state estimation,
which fuses asynchronous observations obtained from an arbitrary number of robots for collaborative localization.
We evaluate our approach in two collaborative object localization scenarios in simulations and on real robots.
Experimental results show that our approach outperforms previous methods and achieves state-of-the-art performance on asynchronous collaborative localization.
\end{abstract}

\section{Introduction}
Object localization is an important area in robotics due to its necessity for improving situational awareness for robots.
It aims to estimate the real-world locations of the objects moving in a dynamic environment
using observations that are acquired by robot sensors such as cameras.
Object localization is widely deployed in a variety of robotics applications, such as autonomous driving to perceive street objects
and search and rescue to find victims.
It is also applied as a critical component in robot capabilities
such as scene reconstruction \cite{delmerico2018comparison,vasquez2014view} and simultaneous localization and mapping (SLAM) \cite{bowman2017probabilistic,sharma2018beyond}.
\begin{figure}
\centering
\vspace{8pt}
\includegraphics[width=0.485\textwidth]{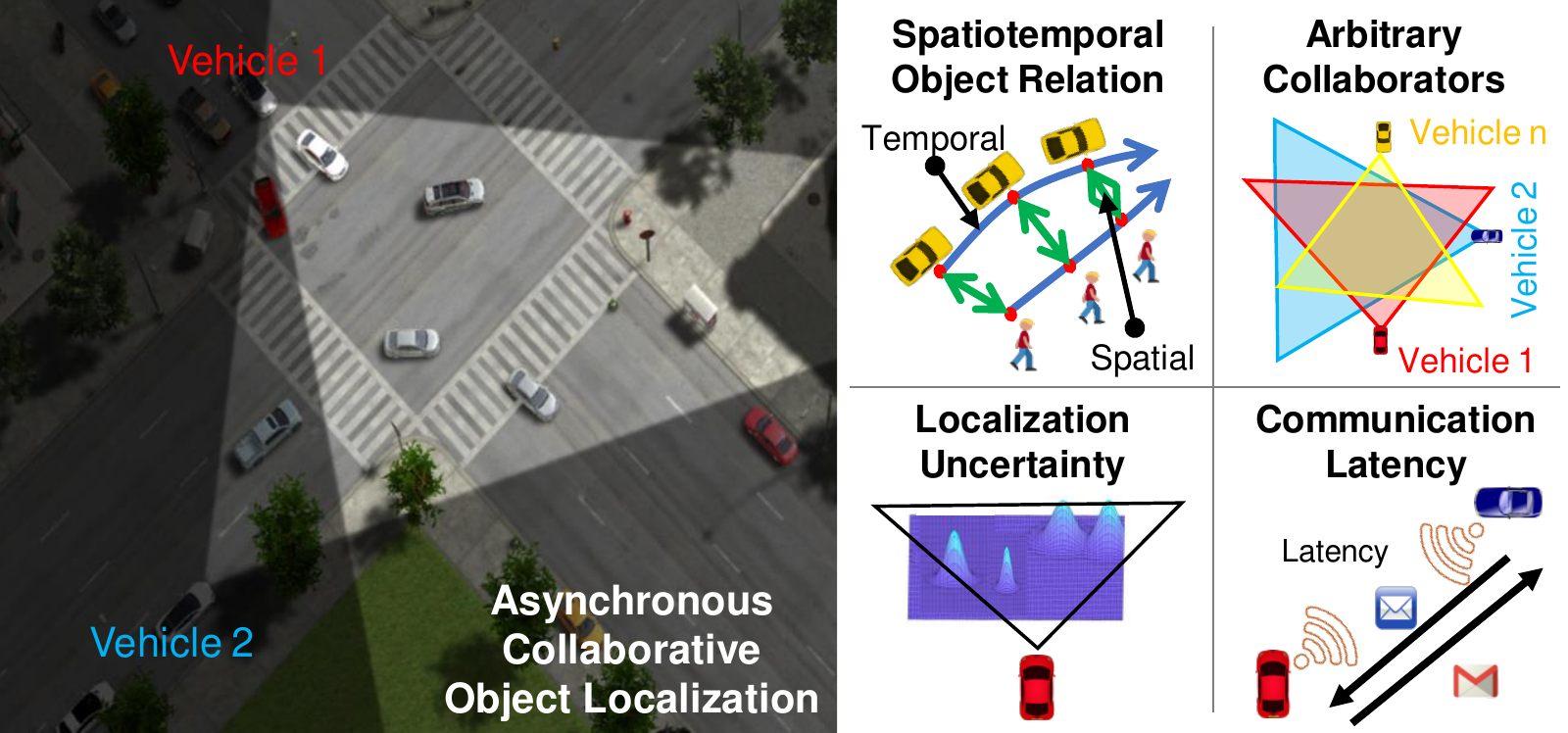}
\caption{A motivating scenario of asynchronous collaborative localization of street objects in connected driving.
When some connected vehicles meet at an intersection, they can improve their shared situational awareness and decrease blind spots by incorporating their observations to collaboratively localize street objects.
}
\label{fig:motivation}
\vspace{-10pt}
\end{figure}
Recently, collaborative object localization using a team of robots has attracted an increased interest because of its improved object localization accuracy and resilience to sensor failures \cite{wei2018survey}.
The goal of collaborative object localization is to estimate locations of observed objects
by fusing observations obtained by multiple robots from different perspectives \cite{acevedo2020dynamic,guo2019collaborative,marvasti2020cooperative,wang2018master}.
For example, as shown in Figure \ref{fig:motivation},
two connected vehicles are able to improve shared situational awareness and decrease blind spots at an intersection
by combining their observations to collaboratively localize street objects.


Given its importance, several approaches have been developed to address collaborative object localization.
Previous methods can be generally grouped into model-based and learning-based methods.
Model-based methods are often based upon Bayesian filtering to localize objects, e.g., using Kalman filters \cite{Weng2020_AB3DMOT} and other Bayesian filters \cite{ullah2017hierarchical}.
These methods lack the ability to model the complex spatiotemporal relationship among objects, e.g., to model the impact of surrounding objects.
The second category of methods use machine learning such as deep neural networks to localize objects \cite{huang2019stgat, ivanovic2019trajectron}.
However, learning-based methods including deep networks assume a fixed number of observations,
and cannot fuse observations from an arbitrary and dynamically changing number of robots.
Moreover, for both categories of approaches, asynchronous observations caused by robot communication delays have not been well addressed.

In this paper, we introduce a novel asynchronous collaborative object localization approach,
which integrates deep spatiotemporal graph learning and Bayesian modeling 
to perform multi-robot sensor fusion in an asynchronous fashion for collaborative object localization.
We encode each observation obtained by each robot in a team as a graph,
where the nodes denote detected objects in the robot's field of view and the edges denote their spatial relationships.
We then encode a sequence of historical observations obtained by each robot as a spatiotemporal graph in order to encode temporal motions of the objects.
Given the representation of spatiotemporal graphs, 
we formulate collaborative object localization as a multi-robot sensor fusion problem for state estimation.
We propose an uncertainty-aware graph learning method to estimate locations of objects the observed by each robot and provide uncertainty quantification.
Then, we introduce a method to integrate spatiotemporal graph learning and model-based estimation for asynchronous collaborative object localization.

The key contribution of the paper focuses on the proposal of an asynchronous collaborative object localization approach.
The novelty of the proposed approach is twofold:
\begin{itemize}
    \item We introduce a novel uncertainty-aware spatiotemporal graph learning approach,
    which is able to represent complex spatiotemporal relationships among observed objects and provide probabilistic estimations of object locations with uncertainty quantification.
    \item We propose a novel method for integrating deep spatiotemporal graph learning and model-based estimation,
    which fuses the asynchronous observations acquired by a team of robots.
    This approach explicitly addresses the latency of robot communications by asynchronous sensor fusion.
    It also offers the ability to fuse observations from an arbitrary and dynamically changing number of robots.
\end{itemize}

\section{Related Work}

Existing techniques for collaborative localization through sensor fusion can be grouped into two categories.
\emph{Model-based approaches} often apply Bayesian filtering to integrate observations obtained from multiple sensors,
e.g., based on Kalman filters \cite{Weng2020_AB3DMOT}, particle filters \cite{deng2019poserbpf,qin2019surgical}, covariance intersection \cite{julier2007using} and Bayesian sequential filters \cite{stenger2006model,ullah2017hierarchical}.
Previous model-based approaches are not able to represent the complex relationships among the objects
or learn from data to improve the localization accuracy.
\emph{Learning-based approaches} apply machine learning methods to fuse multiple observations for object localization.
For example, recurrent neural networks were used to encode object motions and localize the objects \cite{huang2019stgat},
graph neural networks were designed to represent spatial relationships of multiple objects \cite{ivanovic2019trajectron},
and convolutional neural networks were used to model visual-spatial relationships of the objects \cite{chen2017multi,meng2019signet,yin2018geonet}.
Learning-based methods are often not able to integrate observations that are obtained by
an arbitrary and dynamically changing number of robots.


In addition, several learning-based methods were implemented to quantify the uncertainty in object localization, including Bayesian and non-Bayesian methods. Bayesian methods focus on modeling the distribution of model weights, such as Bayes by Backprop \cite{bao2020uncertainty}
and Monte Carlo Dropout \cite{gal2016dropout}. However, Bayesian methods generally  are computationally expensive compared with non-Bayesian methods. A state-of-the-art non-Bayesian uncertainty quantification technique is deep ensemble, which averages predictions from multiple parallel networks to capture the deep network uncertainty \cite{fort2019deep,lakshminarayanan2017simple}. We propose a new multivariate loss function under this framework to quantify the uncertainty of estimated object locations in this paper.

Recently, several hybrid approaches were proposed for state estimation by integrating model-based and learning-based methods,
e.g., by integrating convolutional neural networks and Bayesian filters for robot pose estimation \cite{akai2020hybrid, gao2020multi}, and hybrid sensor fusion based on Gaussian process for location query \cite{barfoot2014batch}.
However, existing hybrid approaches 
cannot well address asynchronous observations caused by the delay of robot communications.
In this paper, we propose one of the first hybrid approaches for asynchronous collaborative object localization that is capable of quantifying localization uncertainty, addressing the latency of robot communications, and addressing the shortcomings of learning and model-based methods.

\section{Approach}\label{sec:approach}

\textbf{Notation.}
We use superscript $t$ and $n$ to denote the time step and the robot index, respectively.
We use subscript $i$ to denote the index of the object in an observation observed by a robot.
For example, $\vv^{t,n}_i$ denotes a feature vector of the $i$-th object observed by the $n$-th robot at time $t$.


\subsection{Problem Formulation}



Given $M$ objects observed by a team of $N$ robots,
we represent each observation obtained by the $n$-th robot as a graph $\GGG^{n}=\{\ZZZ^{n},\RU^{n},\EEE^{n}\},n=1,2,\dots,N$. The node set $\ZZZ^{n}=\{\zz_{1}^{n},\zz_{2}^{n},\dots,\zz_{M}^{n}\}$ denotes the observed locations of objects detected by the $n$-th robot, with $\zz_{i}^{n} \in \RRR^{3}$ denoting the 3D central location of the $i$-th object.
$\RU^{n}=\{\RR_{1}^{n},\RR_{2}^{n},\dots,\RR_{M}^{n}\}$ denotes the uncertainty in observations,
where $\RR_i^{n} \in \RRR^{3 \times 3}$ is defined as the covariance of $\zz_i^{n}$.
The observation $\zz_{i}^{n}$ and its uncertainty $\RR_i^{n}$ can be assumed to follow a multivariate Gaussian distribution $\NNN(\zz_i^{n},\RR_i^{n})$ with $\zz_i^{n}$ as the mean and $\RR_i^{n}$ as the covariance.
The edge set $\EEE^{n}=\{\ee^{n}_{i,j}\}$ denotes the spatial relationships between a pair of detected objects, where $\ee^{t,v}_{i,j}=1$, if $\zz_i^{t,v}$ and $\zz_j^{t,v}$ are connected.
We further encode a sequence of observations recorded from time $1$ to time $t$ by the $n$-th robot as a spatiotemporal graph
$\MMM^{n}=\{\GGG^{1,n},\GGG^{2,n},\dots,\GGG^{t,n}\}$.

We represent the estimated locations of the objects observed by the $n$-th robot at time $t$ as the states $\XXX^{t,n}=\{\xx_i^{t,n}\},i=1,2,\dots M$,
where $\xx_i^{t,n}$ denotes the state estimation (location) of the $i$-th object.
In addition, we use $\PPP^{t,n}=\{\PP_i^{t,n}\},i=1,2,\dots,M$ to denote the uncertainties in $\XXX^{t,n}$, where $\PP_i^{t,n} \in \RRR^{3\times 3}$ denotes the state estimation uncertainty in $\xx_i^{t,n}$.
Then, $\xx_i^{t,n}$ and $\PP_i^{t,n}$ can be assumed to follow the multivariate Gaussian distribution $\NNN(\xx_i^{t,n},\PP_i^{t,n})$.


In this paper, we formulate collaborative object localization as a multi-robot state estimation problem,
with the objective of estimating object states $\XXX^{t,n}$ (i.e., 3D locations of the objects)
by fusing sequences of observations obtained from multiple robots $\{\MMM^{n}\},n=1,2,\dots N$.

\subsection{Uncertainty-Aware Spatiotemporal Graph Learning}\label{sec:learning}
Based on the spatiotemporal graph representations, our approach is able to represent complex spatiotemporal relationships among observed objects and estimates learning-based locations of objects based on historical motions of objects.

Our approach consists of three components.
First, temporal motions of the objects are encoded using a long short-term memory (LSTM) encoder \cite{greff2016lstm},
which is able to process sequential data and mitigates the vanishing gradient problem. Second, the spatial relationships of objects are embedded by a graph attention neural network \cite{velivckovic2017graph}, which captures the spatial impacts of each object from an arbitrary number of neighbor objects (e.g., avoiding collision or changing movement direction).  The spatial and temporal embeddings of  each object are defined as follows:
\begin{align}\label{eq:lstm-encode}    
\mm^{t-1,n}_{i}&=\phi\left(\mm^{t-2,n}_{i}, \zz^{t-1,n}_{i}-\zz^{t-2,n}_{i},\WW^e\right)  \\
\st^{t-1,n}_{i}&=\text{ReLu}\left(\sum_{e^{t-1,n}_{i,j}}\alpha_{i,j}^{t-1,n}\WW^a \mm^{t-1,n}_{j}\right)
\end{align}
where $\zz^{t-1,n}_{i}-\zz^{t-2,n}_{i}$ denotes the relative location (motion) of the $i$-th object observed by the $n$-th robot from time $t-2$ to $t-1$, $\phi$ denotes a one-layer LSTM encoder network with a trainable parameter matrix $\WW^e$ that is shared among all objects, and $\mm^{t-1,n}_{i}$ denotes the temporal motion embedding of the $i$-th object. By iteratively running the encoder, we capture the motion of the object from time $1$ to $t-1$.
In addition, $\WW^a$ is the trainable parameter matrix for the attention network, $\alpha_{i,j}^{t-1,n}$ denotes the impact of the $j$-th neighborhood object on the $i$-th object, which is obtained through the graph attention neural network \cite{velivckovic2017graph},  $e^{t-1,n}_{i,j}$ denotes the connection between the $i$-th object and its $j$-th neighborhood object, $ReLu$ denotes the nonlinear activation function \cite{ramachandran2017searching}, and  $\st^{t-1,n}_{i}$ denotes the spatial embedding of the $i$-th object, which captures the impacts on the $i$-th object from its neighbors.
Given the temporal embedding $\mm^{t-1,n}_{i}$ and spatial embedding $\st^{t-1,n}_{i}$, we predict the states of objects at current time $t$ through a LSTM decoder \cite{greff2016lstm}, which is defined as
\begin{equation} \label{eq:lstm-decoder}
    \xx^{t,n}_{i}=\psi\left(\mm^{t-1,n}_{i}||\st^{t-1,n}_{i}, \zz^{t-1,n}_{i}-\zz^{t-2,n}_{i}, \WW^d\right)
\end{equation}
where $||$ denotes the concatenation operator, $\psi$ denotes a one-layer LSTM decoder network with a trainable parameter matrix $\WW^d$, and $\xx_i^{t,n}$ is the estimated states of the $i$-th object observed by the $n$-th robot at time $t$. By running Eq. (\ref{eq:lstm-decoder}) for multiple times, we can predict arbitrary steps in the future.

Due to the limited amount of training data and noisy observations, the learning-based state estimations obtained from the spatiotemporal graph neural network exhibit uncertainties caused by model bias and perception uncertainty \cite{gal2016dropout}. Inspired by deep ensemble technique \cite{lakshminarayanan2016simple}, we quantify the estimation uncertainties in two steps. 

First, to quantify the data uncertainty, we modify the graph learning model to output the estimated location and the estimation uncertainty in the final layer. For simplification, we write ${\xx}_i^{t,n}$ as ${\xx}^{t}$ by ignoring the index of objects and robots.
The output is assumed to follow the Gaussian distribution, which is defined as 
\begin{equation}\label{eq:gass}
q_{\theta}({\xx}^{t}|\MMM^{1:t-1}) \sim \NNN(\boldsymbol{\mu}_q^t,\boldsymbol{\Sigma}_q^{t})    
\end{equation}
where the mean value $\boldsymbol{\mu}_q^t$ denotes the estimated location, the covariance $\boldsymbol{\Sigma}_q^{t}$ denotes the quantified uncertainty of $\boldsymbol{\mu}_q^t$, and $\theta$ denotes the network parameters. 
To train this model to provide uncertainty estimations, based on the score rule for deep ensemble \cite{lakshminarayanan2016simple}, we design a multivariate negative log-likelihood loss (NLL) defined as follows:
\begin{align}\label{eq:nll}
    L(\theta)&=-\log\left(q_{\theta}\left({\xx}^{t}|\MMM^{1:t-1}\right)\right) \nonumber\\
    &=-\log\left(\frac{1}{\sqrt{2\pi}^3|\boldsymbol{\Sigma}_q^{t}|} e^{(\boldsymbol{\mu}_q^t-\yy^{t})^\top\left(\boldsymbol{\Sigma}^{t}_q\right)^{-1}(\boldsymbol{\mu}_q^t-\yy^{t})}\right) \nonumber\\
    &=\frac{1}{2}(\boldsymbol{\mu}_q^t-\yy^{t})^\top(\boldsymbol{\Sigma}_q^{t})^{-1}(\boldsymbol{\mu}_q^t-\yy^{t}) \nonumber\\
    &\quad +\frac{1}{2}\log(|\boldsymbol{\Sigma}_q^{t}|)+\frac{3}{2}\log(2\pi)
\end{align}
where $|\cdot|$ denotes the determinant operator and $\yy^{t}$ denotes the object's ground truth location at time $t$. $\boldsymbol{\Sigma}_q^{t}$ represents the covariance of the Gaussian distribution with respect to the estimations and the ground truth, which captures the data uncertainty \cite{lakshminarayanan2016simple}. 

Second, to quantify the model uncertainty, 
instead of using just one model, an ensemble of $K$ models is used to form a Gaussian mixture model, which is defined as:
\begin{equation}
    p({\xx}^{t}|\MMM^{1:t-1})
    =\frac{1}{K}\sum_{k=1}^{K}q_{\theta_k}({\xx}^{t}|\MMM^{1:t-1})
\end{equation}
where $\{\theta_k\}^K$ denotes the set of parameters in $K$ networks. Based on deep ensemble theory \cite{lakshminarayanan2016simple}, the Gaussian mixture model can be approximated as the Gaussian distribution, defined as
\begin{equation}\label{eq:est}
    \NNN(\boldsymbol{\mu}_p^t,\boldsymbol{\Sigma}_p^{t})\approx \frac{1}{K}\sum_{k=1}^{K}\NNN(\boldsymbol{\mu}_{q_k}^t,\boldsymbol{\Sigma}_{q_k}^{t})
\end{equation}
Thus, the final estimation is $\xx^{t}=\boldsymbol{\mu}_p^t$ and the estimation uncertainty is defined as:

{\small
\vspace{-6pt}
\begin{align}\label{eq:quantify_Q}
    \QQ^{t}&= \overbrace{\frac{1}{K}\sum_k^K\left(\boldsymbol{\Sigma}_{q_k}^{t}\right)}^{\text{Data uncertainty}}\nonumber\\ &+\overbrace{\frac{1}{K}\sum_k^K\left(\text{diag}(\boldsymbol{\mu}_{q_k}^t)\text{diag}(\boldsymbol{\mu}_{q_k}^t)\right)-\text{diag}(\boldsymbol{\mu}_{p}^t)\text{diag}(\boldsymbol{\mu}_{p}^t)}^{\text{Model uncertainty}}
\end{align}
}\noindent
where $\text{diag}(\cdot)$ is the diagonal operator, and $\QQ_i^{t,n}$ is the process uncertainty, which captures the model and data uncertainties in the learning-based estimation. The data uncertainty is captured by each ensemble model trained with NLL, which describes the ambiguity in targets $\xx^t$ given the inputs $\MMM^{1:t-1}$. The model uncertainty is captured by averaging a combination of ensemble models with consistent training data. If the training data is infinite, then $\boldsymbol{\mu}_{p}^t=\boldsymbol{\mu}_{q_k}^t$. In this case, the model uncertainty will be eliminated and the total uncertainty $\QQ^{t}$ will be reduced to the data uncertainty. 


\subsection{Single-Robot State Estimation}
Based on the quantified process uncertainties, we further integrate learning-based state estimations with single-robot measurements to obtain the the single-robot state estimation.
In conventional sensor fusion methods, the process uncertainties are generally identified through empirical settings \cite{Weng2020_AB3DMOT,gao2020multi}. In this paper, we integrate the quantified process uncertainty into the sensor fusion process as: 
\begin{equation}
  \PP_i^{t,n}= \PP_i^{t-1,n} + \QQ_i^{t,n}  
\end{equation}
where $\PP_i^{t,n}$ denotes the uncertainty of the learning-based state estimation $\xx^{t,n}_{i}$, which is computed by the sum of the process uncertainty $\QQ_i^{t,n}$ and the state estimation uncertainty at the previous time step. Based on Kalman filter \cite{li2015kalman}, we update the learning-based state estimation with the measurement $\zz^{t,n}$ to obtain the single-robot state estimation, which is defined as follows:
\begin{align}\label{eq:single-est}
    \KK^{t,n}&=\PP_i^{t,n}(\PP_i^{t,n}+\RR_i^{t,n})^{-1}  \\
    \hat{\xx}_i^{t,n}&=\xx_i^{t,n}+\KK^{t,n}(\zz_i^{t,n}-\xx_i^{t,n})   \\
     \hat{\PP}_i^{t,n}&=(\II-\KK_i^{t,n}) \PP_i^{t,n}
\end{align}
where $\zz^t$ denotes the measurement at time $t$, $\RR_i^{t,n}$ denotes the uncertainty in the measurement, $ \KK^{t,n}$ denotes the Kalman gain that encodes the relative weight of both the state estimation and the measurement given their uncertainties, $\hat{\xx}_i^{t,n}$ denotes the updated state estimation and $\hat{\PP}_i^{t,n}$ denotes the updated uncertainty of the single-robot state estimation.
\subsection{Asynchronous Multi-robot State Estimation}
Based on the single-robot state estimation and the state estimation uncertainty, our approach computes a final estimation for each robot by integrating multi-robot state estimations. We also address the challenges of asynchronous multi-robot state estimations caused by communication time delays and integrating arbitrary number of state estimations provided by multiple robots.

In real-world multi-robot collaborative object localization, asynchronous multi-robot state estimations caused by communication time delays usually significantly increase the error in object localization, as the time-delayed locations of objects can be far away from the current locations of the objects.
In order to address asynchronous state estimations, we first compensate the state estimations with time delay by using a deep graph learning network as follows:
\begin{equation}\label{eq:cmp_est}
    \tilde{\xx}_i^{t,n},\tilde{\QQ}_i^{t,n}=\Phi\left(\hat{\xx}_i^{1,n},\hat{\xx}_i^{2,n},\dots,\hat{\xx}_i^{t-\Delta T,n}\right)
\end{equation}
where $ \tilde{\xx}_i^{t,n}$ and estimation uncertainty $\tilde{\QQ}_i^{t,n}$ denotes the compensated state estimation and estimation uncertainty.
$\Phi$ denotes the deep graph learning network, which has the same architecture as the the uncertainty-aware graph neural network. 
$\Delta T$ denotes the time delay in the state estimation provided by the $n$-th robot, which can be obtained through timestamps or learning models \cite{zhang2020single}. 

Since the compensation process also introduces uncertainty into the state estimation, we update the uncertainty of the compensated state estimation as follows:
\begin{equation}\label{eq:update_cmp}
    \tilde{\PP}_i^{t,n}= \hat{\PP}_i^{t-\Delta T,n}+ \tilde{\QQ}_i^{t,n}
\end{equation}
where $\tilde{\PP}_i^{t,n}$ denotes the updated uncertainty of the compensated state estimation, which is obtained by the sum of the process uncertainties $\tilde{\QQ}_i^{t,n}$ generated in the compensation process and the state estimation uncertainty $\hat{\PP}_i^{t-\Delta T,n}$ computed $\Delta T$ time ago.
To improve the robustness to noise in collaborative object localization, we propose a multi-robot fusion gain to integrate state estimations provided by an arbitrary number of collaborative robots, which is defined as follows:
\begin{equation}\label{eq:fusion gain}
    \EE_i^{t,n}=\left(\sum_{j=1}^{N} (\tilde{\PP}_i^{t,j})^{-1}\right)^{-1}(\tilde{\PP}_i^{t,n})^{-1}
\end{equation}
where $\EE_i^{t,n} \in \RRR^{3\times 3}$ denotes the state fusion gain for each robot's state estimations.
In addition, $\EE_i^{t,n}$ follows the constraint $\sum_{n=1}^{N} \EE_i^{t,n}=\mathbf{I}$, where $\mathbf{I} \in \RRR^{3\times 3}$ denotes an identity matrix.  The fusion gain for each robot represents the weight of each robot's state estimation in all the multi-robot state estimations given the normalized state estimation uncertainties.
The final state is defined as follows:
\begin{equation} \label{eq:multi_update_state}
    \tilde{\xx}_i^{\prime t,n}= \EE_i^{t,n}\tilde{\xx}_i^{t,n}+ \sum_{j=1,j\ne n}^{N}\EE_i^{t,j}\sigma(\tilde{\xx}_i^{t,j})
\end{equation}
where $\sigma$ denotes the transformation function that transforms the multi-robot state estimations to the $n$-th robot's coordinates based on camera extrinsic parameters \cite{zhang2004extrinsic}. The camera extrinsic parameters can be obtained through GPS \cite{brahmbhatt2018geometry} or deep learning algorithm \cite{kendall2015posenet}. $\tilde{\xx}_i^{\prime t,n}$ denotes the final state estimation of the $i$-th object observed by the $n$-th robot at time $t$, which is computed by the sum of single-robot state estimations weighted by the fusion gains. If a robot's state estimation has large uncertainty (e.g., existing large communication delay), then its contribution will be heavily weakened during the fusion. The uncertainty of the final state estimation is defined as 
\begin{equation} \label{eq:multi_update_un}
    \tilde{\PP}_i^{\prime t,n}=\left( \sum_{n=1}^N (\tilde{\PP}_i^{t,n})^{-1}\right)^{-1}
\end{equation}
where $\tilde{\PP}_i^{\prime t,n}$ denotes the uncertainty of the final state estimation $\tilde{\xx}_i^{\prime t,n}$, which is obtained by integrating all the single-robot state estimation uncertainties.

The complexity of our algorithm is $O(NKG)$, where $O(G)$ denotes the complexity of the graph learning network. Since the time compensation for each robot can be run parallel, the complexity of our algorithm reduces to $O(KG)$. Since the ensemble models can also be run parallel, the complexity can further reduce to $O(G)$.
When our algorithm is executed on a Linux machine with an i9 3.6 GHz CPU, 3 TB memory, and an Nvidia RXT 2080Ti GPU, the average runtime is around 10 Hz.

\begin{figure}[ht]
\centering
\vspace{6pt}
\subfigure[CAD]{
\centering
\includegraphics[width=0.48\textwidth]{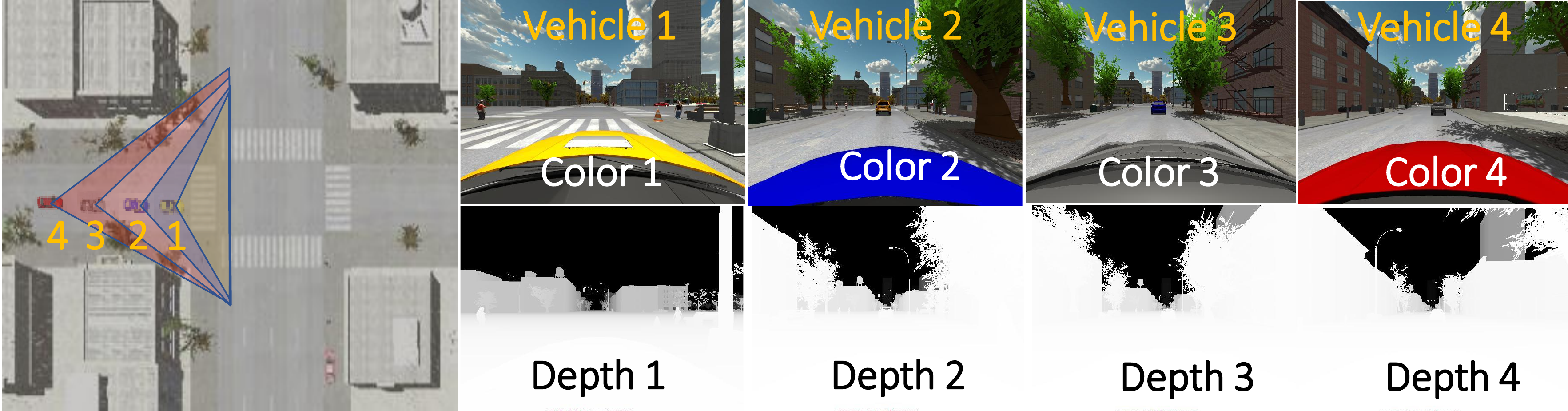}\label{fig:cad}
}\\
\vspace{-4pt}
\subfigure[MROL]{
\centering
\includegraphics[width=0.48\textwidth]{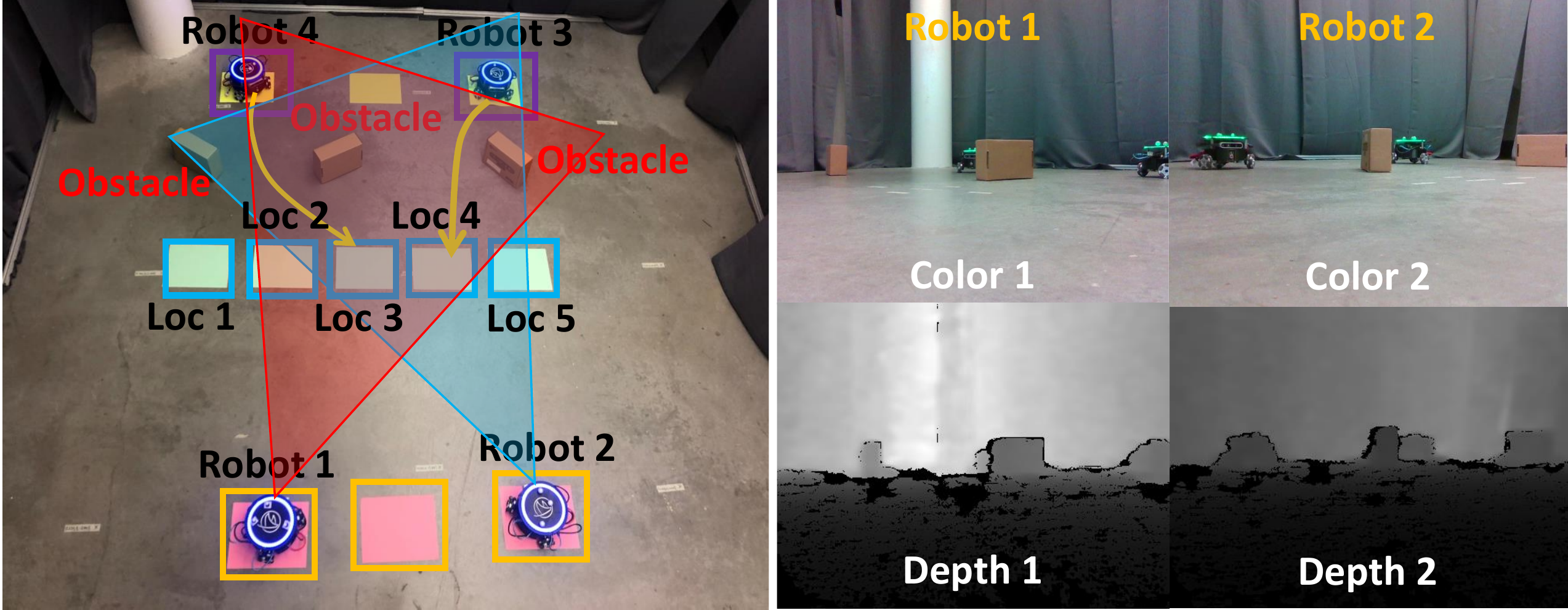}\label{fig:mrta}
}
\vspace{-8pt}
\caption{Illustrations of the simulated Connected Autonomous Driving (CAD) and Multi-Robot Object Localization (MROL) scenarios that are used in our experiments.
}\label{fig:dataset}
\vspace{-6pt}
\end{figure}
\section{Experiments}

\subsection{Experimental Setup}


We use both high-fidelity robotic simulations and physical robots to evaluate our method for asynchronous collaborative object localization in multi-robot systems. Our evaluation consists of two scenarios, including simulated Connected Autonomous Driving (CAD) and real-world Multi-Robot Object Localization (MROL), as demonstrated in Figure \ref{fig:dataset}. 
CAD includes $1500$ data instances recorded at $10$Hz by four RGB-D cameras mounted on the autonomous vehicles. The ground truth locations are obtained from the connected vehicle simulator provided by Toyota.
MROL includes $800$ data instances, which are recorded by two RGB-D cameras mounted on robots at $10$Hz. The ground truth locations are obtained from an Optitrack motion tracking system.

In the experiments, we only perform our approach to the objects observed by more than one robot and the association of the same object in different observations is identified based on the recent work \cite{gaoregularized}. 
The node attributes are generated from 3D locations of objects and the edges are fully connected.
The LSTM encoder in Eq. (\ref{eq:lstm-encode}) and decoder in Eq. (\ref{eq:lstm-decoder}) are constructed by only one LSTM layer, with $\WW^e$ setting to the dimension of $3 \times 32$ and $\WW^d$ setting to $64 \times 6$. The graph attention network is constructed with two layers,  with $\WW^a$ having the dimension of $32 \times 64$ followed by dropout with probability $0.1$.
The number of ensemble models is $K=5$. Initially, the state $\xx$ is set to an all zero matrix and the state estimation uncertainty $\PP$ is a diagonal matrix with the diagonal values set to $10$. Observation  uncertainty $\RR$ is calculated based on the depth sensor model given the real depth values of objects \cite{khoshelham2011accuracy}. ADMM is used as the optimization solver \cite{boyd2011distributed}.

We implemented our full approach and two baseline methods for asynchronous collaborative object localization.
The first baseline method named \textbf{Ours-U} that only quantifies the process uncertainty without addressing asynchronous observations caused by communication latency among multiple robots.
The second baseline method named \textbf{Ours-D} that only addresses asynchronization and uses a fixed value as the process uncertainty.
In addition, we compare with four previous methods, including
\textbf{AOM} \cite{ji2017surfacenet} that directly averages the locations of the same objects in different observations,
\textbf{MMT} \cite{Weng2020_AB3DMOT} that estimates the locations of objects based on the classical Kalman filter,
\textbf{STTP} \cite{huang2019stgat} that localizes objects based on spatiotemporal graph learning,
\textbf{MCOL} \cite{gao2020multi} that integrates model-based and learning-based estimations for object localization, but without considering uncertainty quantification and asynchronous sensing data.

We follow a widely used experimental setup \cite{huang2019stgat,godard2017unsupervised} to evaluate our approach. Displacement error (\textbf{DE}) is used to evaluate the localization accuracy, which is defined as the Euclidean distance between the estimated location and the ground truth location. The unit of DE is meter. Relative displacement error (\textbf{REL-DE}) is used to evaluate the localization accuracy relative to the measurement distance, which is defined as the ratio of the displacement error over the ground truth location.

\subsection{Connected Autonomous Driving Simulation}

\begin{table}
\centering
\vspace{6pt}
\tabcolsep=0.3cm
\caption{Quantitative results on CAD and MROL.}
\label{tab:QuanResults}
\begin{tabular}{|l|c|c|c|c|}
\hline
Method & \multicolumn{2}{ c| }{CAD}  & \multicolumn{2}{ c| }{MROL}  \\
\cline{2-5}
		 & DE & Rel-DE & DE  & Rel-DE\\
		\hline\hline
		\hline
		AOM \cite{ji2017surfacenet}   &2.2288 & 0.1197  &0.0343  & 0.0283      \\
		\hline
		STTP \cite{huang2019stgat} &1.7859   & 0.0950  &0.0357 & 0.0294       \\
		\hline
		MMT \cite{Weng2020_AB3DMOT} &1.6240   & 0.0866     &0.0344 & 0.0283  \\
		\hline
		MCOL \cite{gao2020multi} &1.4445   &0.0771    &0.0319 & 0.0263    \\
		\hline\hline
		\textbf{Ours-D}  &1.4397 &0.0768           &0.0310  &0.0256     \\
		\textbf{Ours-U}  &1.3688 &0.0730  &0.0304  & 0.0251  \\
		\textbf{Ours}   &\textbf{1.3533} &\textbf{0.0723}      &\textbf{0.0288}  & \textbf{0.0238}    \\
		\hline	
\end{tabular}
\vspace{-6pt}
\end{table}
Our approach is first evaluated on the CAD scenario, aiming to localize the dynamic pedestrians and vehicles given the observations from multiple connected vehicles. The number of connected vehicles is from $1$ to $4$, and communication delays between connected vehicles is from $0.1$ to $0.7$ sec.

The qualitative results over CAD are presented in Figures \ref{fig:cad-mmt}-\ref{fig:cad-ours}. We observe that the movement of the yellow vehicle has a sharp turn, which is challenging to accurately localize it. In this situation, our approach works much better compared with the other two methods. 
Moreover, we observe that MMT works poorly as shown in Figure \ref{fig:cad-mmt}, as the linear velocity assumption in MMT cannot estimate the non-linear movement of objects. MCOL improves the localization performance, however, its estimations contain a lagging effect (the estimated locations lag behind the ground truth), which is caused by the asynchronous state estimations provided by multiple connected vehicles.
Our approach outperforms all these methods, due to its ability to model the complex spatiotemporal relationship among objects, and to address multi-robot asynchronous state estimations.

The quantitative results for the CAD scenario are shown in Table \ref{tab:QuanResults}. We observe that our baseline methods outperform the compared previous methods, which indicates the importance of quantifying estimation uncertainty and addressing asynchronous multi-robot state estimations caused by communication latency for collaborative object localization. In addition,
AOM performs poorly due to the noisy observations. STTP and MMT obtain an improved performance by utilizing learning-based state estimation to encode historical estimations, and by using model-based sensor fusion for object localization.  MCOL further improves the performance by integrating learning-based and model-based state estimations.
Due to uncertainty quantification and addressing asynchronous state estimations resulting from communication latency, our approach can readily integrate multi-robot asynchronous state estimations and performs the best.

\subsection{Real-world Multi-robot Object Localization}

Our method is also evaluated in the MROL scenario. In this scenario, multiple robots collaboratively localize the objects to assist with collision avoidance. The object instances used in this scenario include different dynamic robots. The MROL scenario is challenging due to the large noise, large communication latency (ranging from $0.1$ sec to $0.7$ sec) and uncertainty existing in robot observations, which is caused by non-smooth movements of robots and occlusion.

\begin{figure}[ht]
\centering
\subfigure[MMT]{\includegraphics[width=0.15\textwidth]{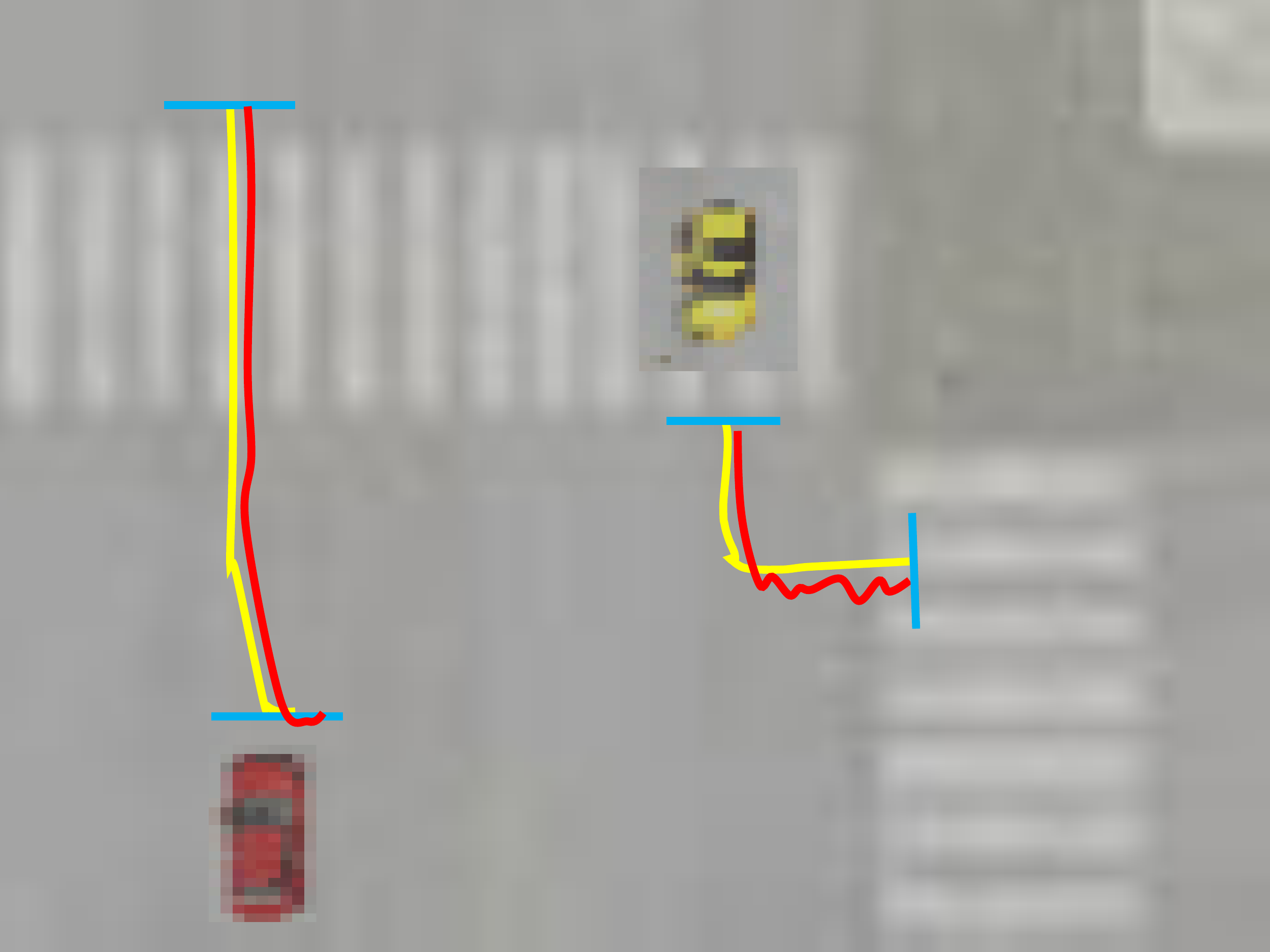}\label{fig:cad-mmt}}
\subfigure[MCOL]{\includegraphics[width=0.15\textwidth]{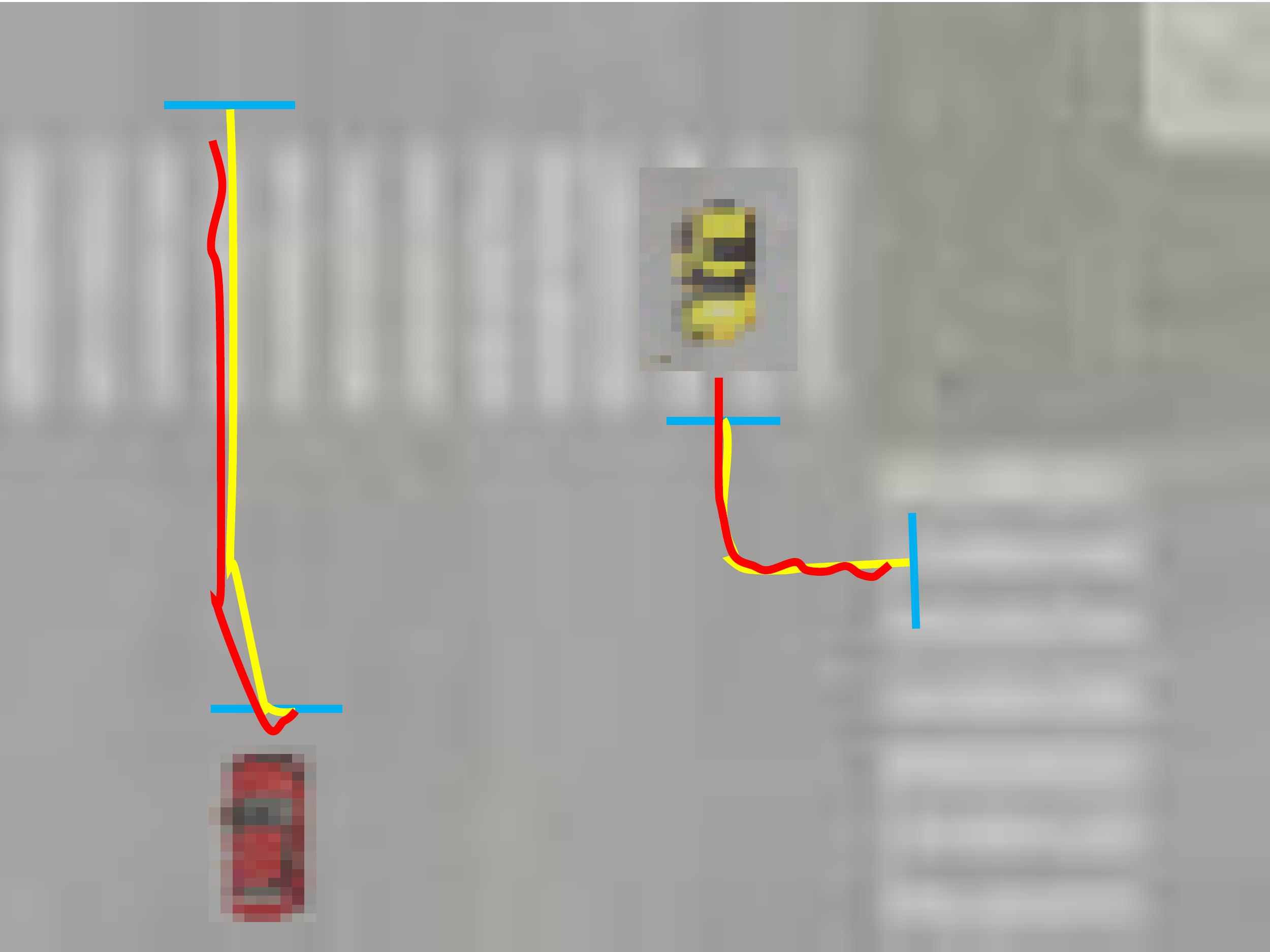}\label{fig:cad-cmol}}
\subfigure[Ours]{\includegraphics[width=0.15\textwidth]{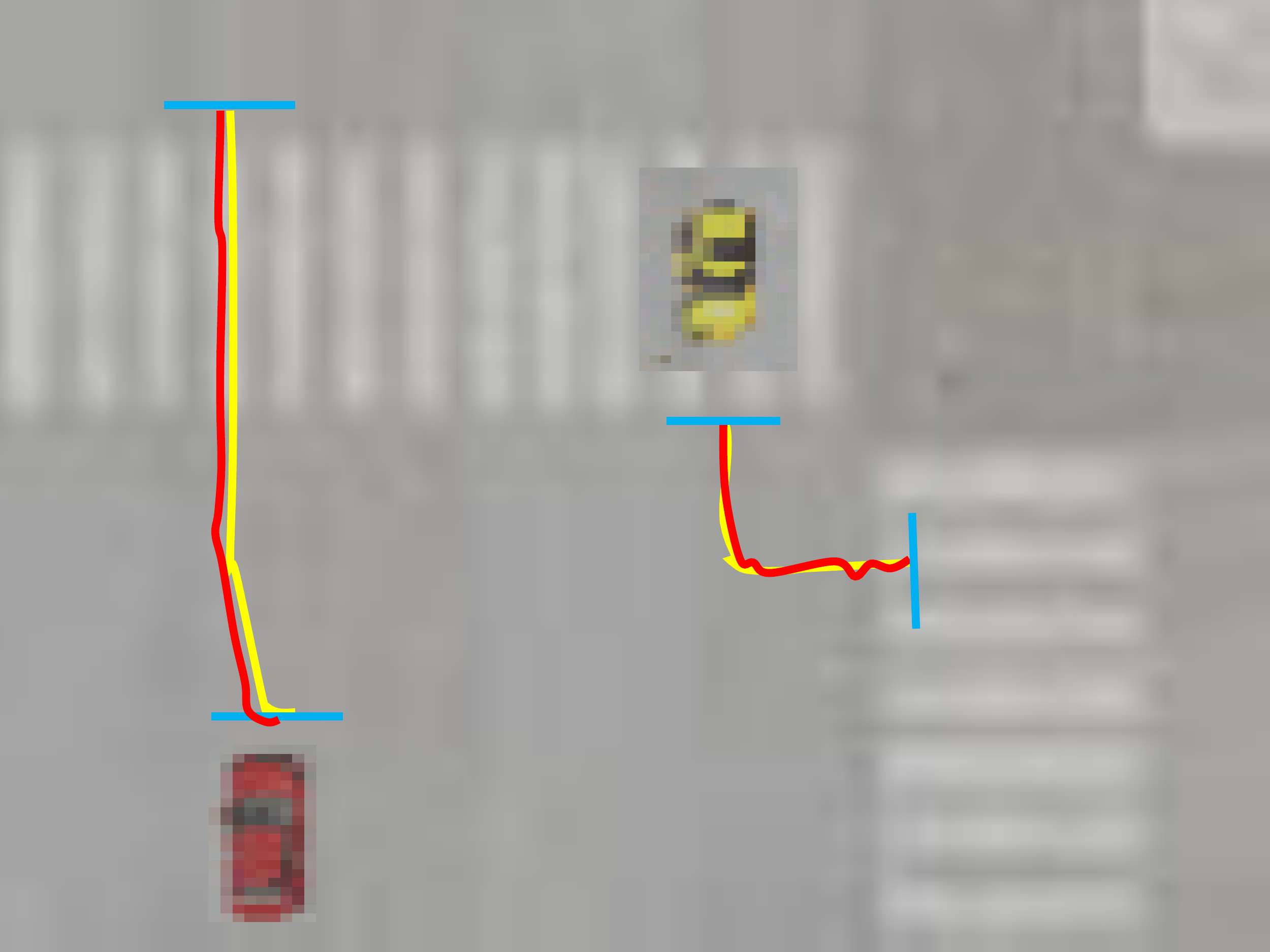}\label{fig:cad-ours}}
\\
\subfigure[MMT]{\includegraphics[width=0.15\textwidth]{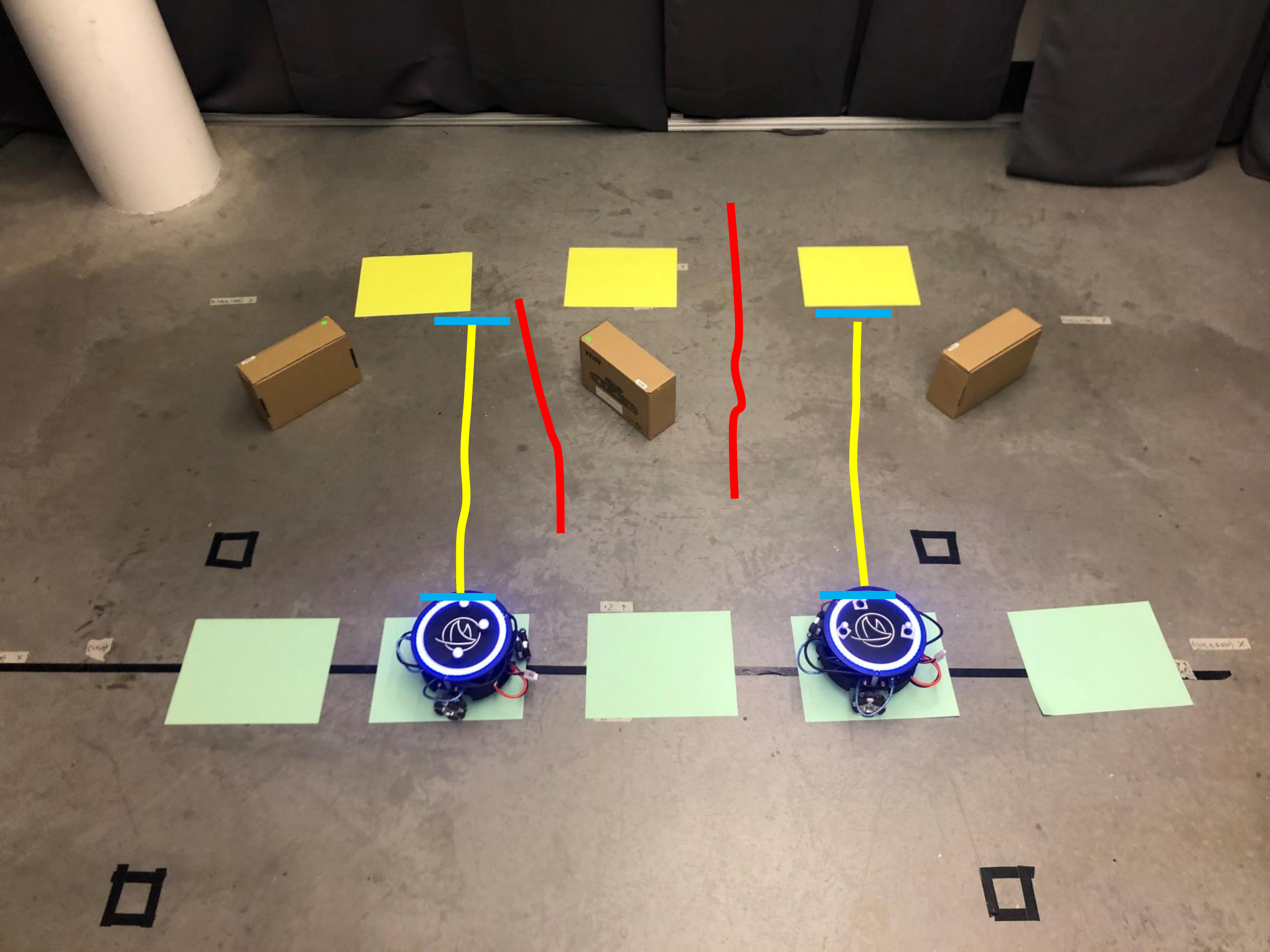}\label{fig:mrta-mmt}}
\subfigure[MCOL]{\includegraphics[width=0.15\textwidth]{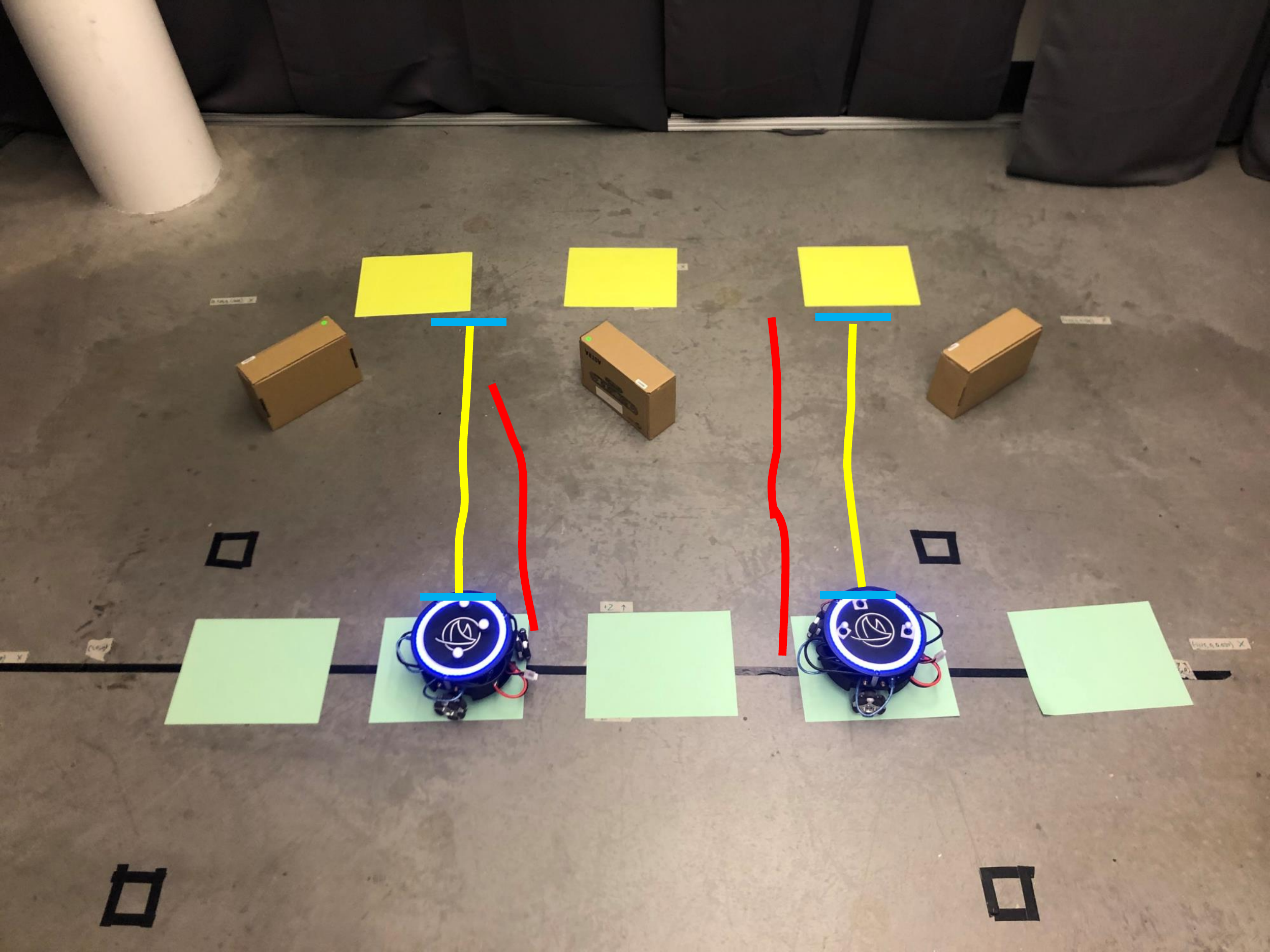}\label{fig:mrta-cmol}}
\subfigure[Ours]{\includegraphics[width=0.15\textwidth]{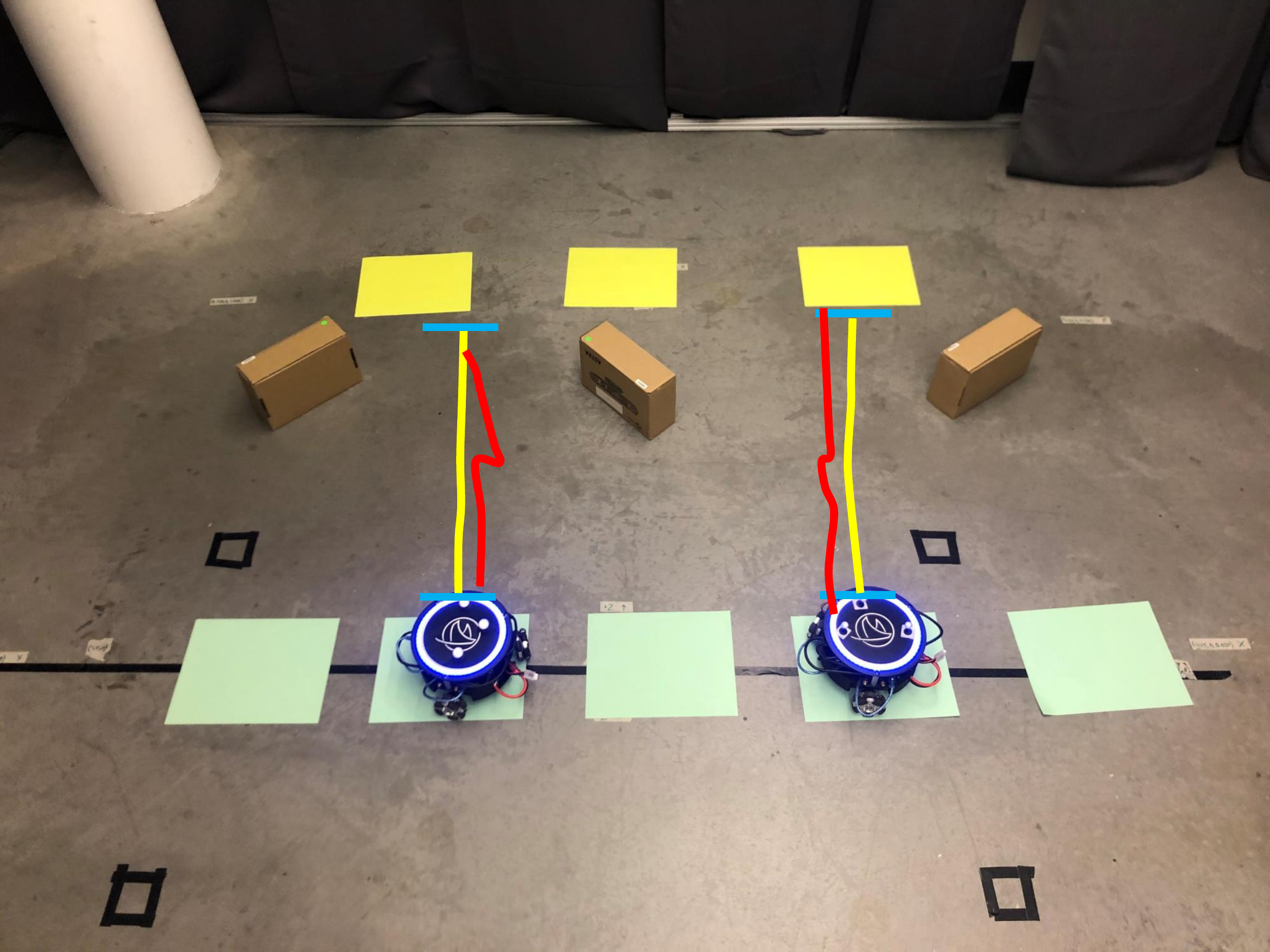}\label{fig:mrta-ours}}
\caption{Qualitative experimental results of our approach and comparisons with the previous and baseline methods on CAD and MROL. Ground truth paths are shown in yellow, the estimated paths are shown in red, and the blue lines denotes the starting and ending points of trajectories (indicating the lagging effects in Figure \ref{fig:cad-cmol} and \ref{fig:mrta-cmol}).
}
\label{fig:QualResults}
\end{figure}

The qualitative results obtained on MROL are presented in Figures \ref{fig:mrta-mmt}-\ref{fig:mrta-ours}. We can see visually that our approach outperforms MMT and MCOL. Specifically,
MCOL's estimations lag the ground truth caused by the communication latency in the multi-robot system. In addition, our approach has better localization accuracy than MCOL, as we use the quantified process uncertainty in the sensor fusion process, which can appropriately weight learning-based state estimations during sensor fusion, rather than using a constant value as in MCOL. Thus, the proposed approach achieves the best performance in this realistic multi-robot object localization scenario.
The quantitative results obtained in the MROL scenario are presented in Table \ref{tab:QuanResults}. Similar to the other scenarios,
our approach and its baseline methods also outperform MCOL by quantifying the uncertainty of the learning process and addressing asynchronous state estimations, resulting in the best performance on this real-world multi-robot scenario.

\begin{figure}
\centering
\vspace{6pt}
\subfigure[Latency]{\includegraphics[height=2.7cm]{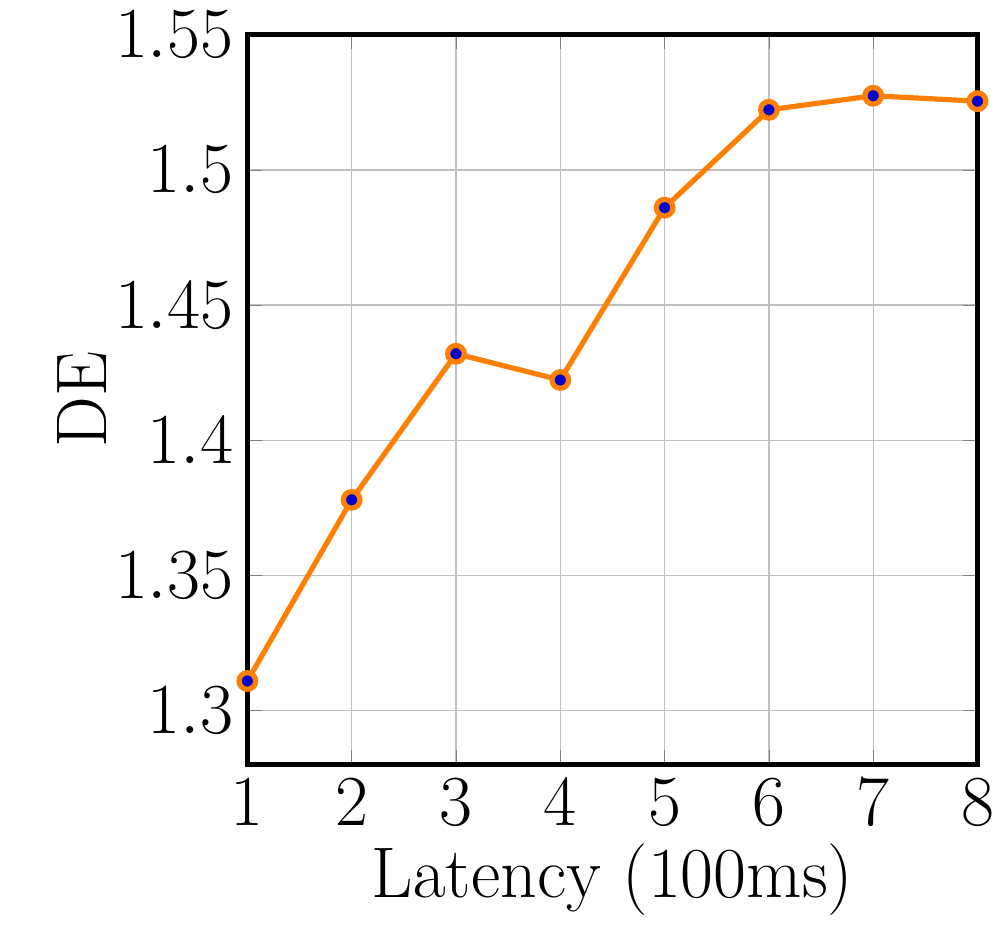}\label{fig:delay}}
\hspace{-8pt}
\subfigure[\# Collab. ]{\includegraphics[height=2.7cm]{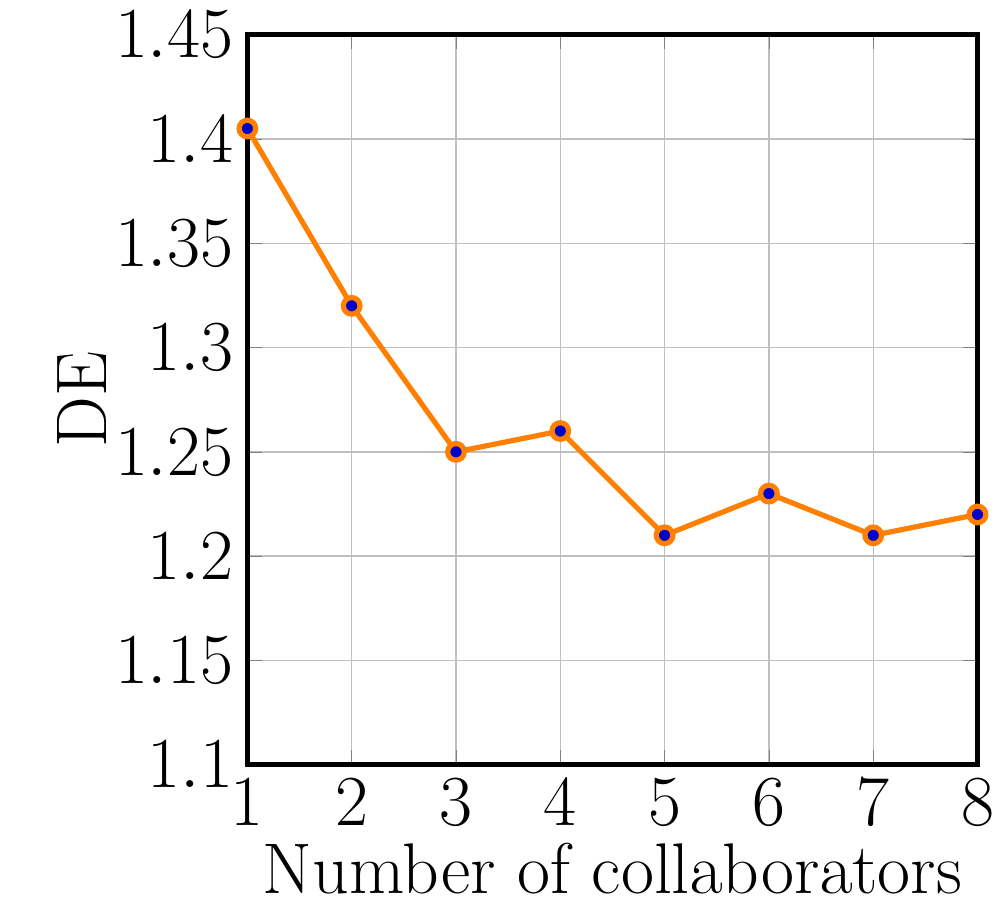}\label{fig:num}}
\hspace{-8pt}
\subfigure[Robustness]{\includegraphics[height=2.7cm]{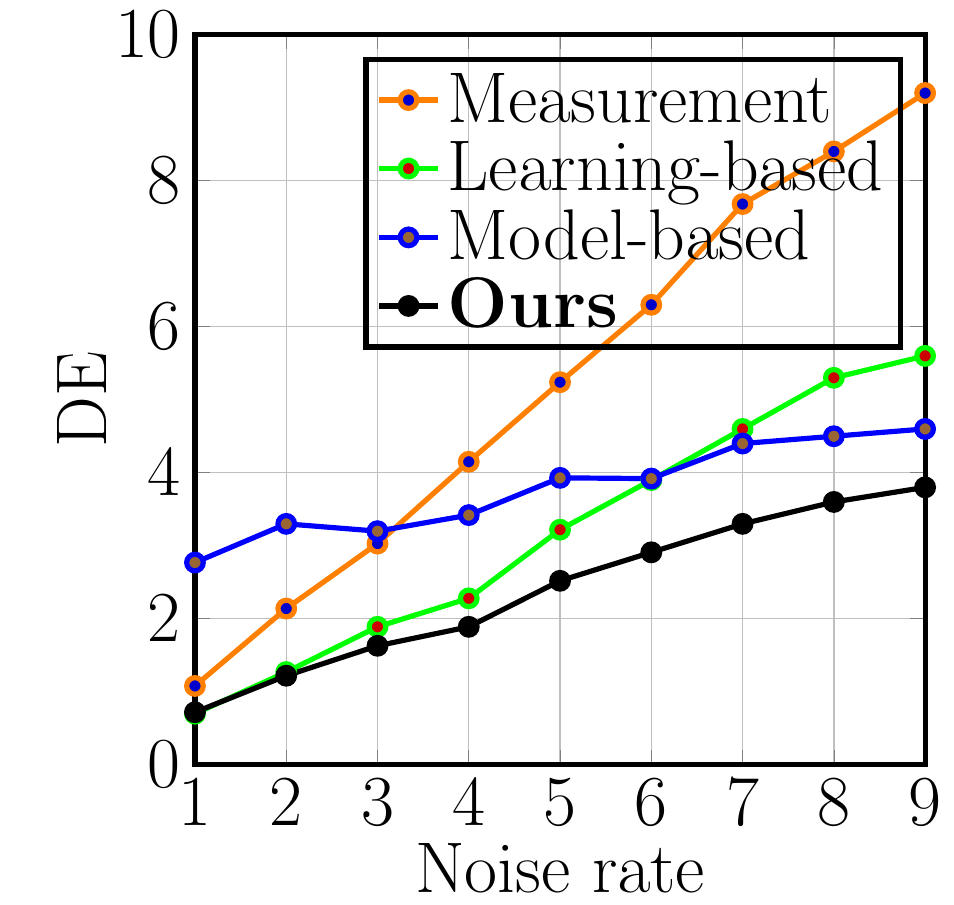}\label{fig:noise}}
\caption{Analysis of our method's characteristics.}
\label{fig:discussion}
\vspace{-9pt}
\end{figure}

\subsection{Discussion}

Using the CAD simulations, we further investigate our approach's characteristics, including the effect of communication latency, the number of collaborating robots, and the robustness to noise.

Figure \ref{fig:delay} depicts how our approach is affected by increasing communication time delays among vehicles, which causes asynchronous state estimations. We observe that the performance of our approach gradually decreases as the communication latency among robots increases. When the communication latency increases beyond $600$ ms, the performance becomes stable with small fluctuations. 
In this case, each robot uses only its own observations to localize objects without using any collaboration from other robots. The figure generally shows that our approach can effectively fuse asynchronous multi-robot state estimations with large communication delays. 


Figure \ref{fig:num} illustrates the influence of the number of collaborative robots on CAD. We can see that the performance of our approach gradually improves as the number of collaborators increases.   When the number of collaborators is larger than $5$, the performance becomes stable with small fluctuations. The best performance is around $[1.2145,1.2373]$. Thus, integrating multi-robot observations can effectively improve the performance of collaborative object localization.


Figure \ref{fig:noise} illustrates the effect of varying noise rates. 
It compares the performance of measurement-only localization, only the learning-based state estimation, only the model-based state estimation, and on our complete method.
We observe that the area under using measurements directly is $0.5068$, the area under only using model-based state estimation is $0.3744$, the area under only using learning-based state estimations is $0.3217$ and our complete approach is $0.2561$. Thus, integrating model and learning-based estimations effectively improves the robustness to noise.

\section{Conclusion}


In this paper, we propose a novel asynchronous collaborative object localization approach that integrates uncertainty-aware spatiotemporal graph learning and model-based state estimation in a principled way to perform asynchronous collaborative object localization. Our approach can learn spatiotemporal relationships among objects to provide probabilistic estimations of object locations. In addition, we propose a new method to integrate learning and model-based state estimations, which is able to fuse asynchronous observations obtained from an arbitrary number of robots to collaboratively localize objects.
Extensive experiments are conducted to evaluate our approach and the experimental results show that our approach outperforms existing methods and achieves state-of-the-art performance on asynchronous collaborative object localization.

\bibliographystyle{IEEEtran}
\bibliography{ref}

\begin{thebibliography}{10}
\providecommand{\url}[1]{#1}
\csname url@rmstyle\endcsname
\providecommand{\newblock}{\relax}
\providecommand{\bibinfo}[2]{#2}
\providecommand\BIBentrySTDinterwordspacing{\spaceskip=0pt\relax}
\providecommand\BIBentryALTinterwordstretchfactor{4}
\providecommand\BIBentryALTinterwordspacing{\spaceskip=\fontdimen2\font plus
\BIBentryALTinterwordstretchfactor\fontdimen3\font minus
  \fontdimen4\font\relax}
\providecommand\BIBforeignlanguage[2]{{%
\expandafter\ifx\csname l@#1\endcsname\relax
\typeout{** WARNING: IEEEtran.bst: No hyphenation pattern has been}%
\typeout{** loaded for the language `#1'. Using the pattern for}%
\typeout{** the default language instead.}%
\else
\language=\csname l@#1\endcsname
\fi
#2}}

\bibitem{delmerico2018comparison}
J.~Delmerico, S.~Isler, R.~Sabzevari, and D.~Scaramuzza, ``{A} comparison of
  volumetric information gain metrics for active {3D} object reconstruction,''
  \emph{AuRo}, vol.~42, no.~2, pp. 197--208, 2018.

\bibitem{vasquez2014view}
J.~I. Vasquez-Gomez, L.~E. Sucar, and R.~Murrieta-Cid, ``{View} planning for
  {3D} object reconstruction with a mobile manipulator robot,'' in \emph{IROS},
  2014.

\bibitem{bowman2017probabilistic}
S.~L. Bowman, N.~Atanasov, K.~Daniilidis, and G.~J. Pappas, ``{Prob}abilistic
  data association for semantic {SLAM},'' in \emph{ICRA}, 2017.

\bibitem{sharma2018beyond}
S.~Sharma, J.~A. Ansari, J.~K. Murthy, and K.~M. Krishna, ``{Beyond} pixels:
  {L}everaging geometry and shape cues for online multi-object tracking,'' in
  \emph{ICRA}, 2018.

\bibitem{wei2018survey}
S.~Wei, D.~Yu, C.~L. Guo, L.~Dan, and W.~W. Shu, ``{Survey} of connected
  automated vehicle perception mode: from autonomy to interaction,''
  \emph{ITS}, vol.~13, no.~3, pp. 495--505, 2018.

\bibitem{acevedo2020dynamic}
J.~J. Acevedo, J.~Messias, J.~Capit{\'a}n, R.~Ventura, L.~Merino, and P.~U.
  Lima, ``{A} dynamic weighted area assignment based on a particle filter for
  active cooperative perception,'' \emph{RAL}, vol.~5, no.~2, pp. 736--743,
  2020.

\bibitem{guo2019collaborative}
R.~Guo, H.~Lu, P.~Gao, Z.~Zhang, and H.~Zhang, ``{C}ollaborative localization
  for occluded objects in connected vehicular platform,'' in \emph{VTC}, 2019.

\bibitem{marvasti2020cooperative}
E.~E. Marvasti, A.~Raftari, A.~E. Marvasti, Y.~P. Fallah, R.~Guo, and H.~Lu,
  ``{Co}operative {Lidar} object detection via feature sharing in deep
  networks,'' \emph{ArXiv}, 2020.

\bibitem{wang2018master}
H.~Wang, C.~Zhang, Y.~Song, and B.~Pang, ``{Ma}ster-followed multiple robots
  cooperation {SLAM} adapted to search and rescue environment,'' \emph{Int J
  Control Autom Syst}, 2018.

\bibitem{Weng2020_AB3DMOT}
X.~Weng, J.~Wang, D.~Held, and K.~Kitani, ``{3D} multi-object tracking: {A}
  baseline and new evaluation metrics,'' \emph{IROS}, 2020.

\bibitem{ullah2017hierarchical}
M.~Ullah, A.~K. Mohammed, F.~A. Cheikh, and Z.~Wang, ``{A} hierarchical feature
  model for multi-target tracking,'' in \emph{ICIP}, 2017.

\bibitem{huang2019stgat}
Y.~Huang, H.~Bi, Z.~Li, T.~Mao, and Z.~Wang, ``{STGAT: Modeling}
  spatial-temporal interactions for human trajectory prediction,'' in
  \emph{ICCV}, 2019.

\bibitem{ivanovic2019trajectron}
B.~Ivanovic and M.~Pavone, ``{The} trajectron: {P}robabilistic multi-agent
  trajectory modeling with dynamic spatiotemporal graphs,'' in \emph{ICCV},
  2019.

\bibitem{deng2019poserbpf}
X.~Deng, A.~Mousavian, Y.~Xiang, F.~Xia, T.~Bretl, and D.~Fox, ``{PoserBPF}:
  {A} rao-blackwellized particle filter for {6D} object pose tracking,''
  \emph{RSS}, 2019.

\bibitem{qin2019surgical}
F.~Qin, Y.~Li, Y.-H. Su, D.~Xu, and B.~Hannaford, ``{S}urgical instrument
  segmentation for endoscopic vision with data fusion of rediction and
  kinematic pose,'' in \emph{ICRA}, 2019.

\bibitem{julier2007using}
S.~J. Julier and J.~K. Uhlmann, ``Using covariance intersection for slam,''
  \emph{Robotics and Autonomous Systems}, vol.~55, no.~1, pp. 3--20, 2007.

\bibitem{stenger2006model}
B.~Stenger, A.~Thayananthan, P.~H. Torr, and R.~Cipolla, ``{Mo}del-based hand
  tracking using a hierarchical bayesian filter,'' \emph{PAMI}, vol.~28, no.~9,
  pp. 1372--1384, 2006.

\bibitem{chen2017multi}
X.~Chen, H.~Ma, J.~Wan, B.~Li, and T.~Xia, ``{M}ulti-view {3D} object detection
  network for autonomous driving,'' in \emph{CVPR}, 2017.

\bibitem{meng2019signet}
Y.~Meng, Y.~Lu, A.~Raj, S.~Sunarjo, R.~Guo, T.~Javidi, G.~Bansal, and
  D.~Bharadia, ``{SIGNET: S}emantic instance aided unsupervised {3D} geometry
  perception,'' in \emph{CVPR}, 2019.

\bibitem{yin2018geonet}
Z.~Yin and J.~Shi, ``{GeoNet}: {Unsupervised} learning of dense depth, optical
  flow and camera pose,'' in \emph{CVPR}, 2018.

\bibitem{bao2020uncertainty}
W.~Bao, Q.~Yu, and Y.~Kong, ``{U}ncertainty-based traffic accident anticipation
  with spatio-temporal relational learning,'' in \emph{ACM-MM.}, 2020.

\bibitem{gal2016dropout}
Y.~Gal and Z.~Ghahramani, ``{D}ropout as a {Bayesian} approximation:
  Representing model uncertainty in deep learning,'' in \emph{ICML}, 2016.

\bibitem{fort2019deep}
S.~Fort, H.~Hu, and B.~Lakshminarayanan, ``{D}eep ensembles: {A} loss landscape
  perspective,'' \emph{ArXiv}, 2019.

\bibitem{lakshminarayanan2017simple}
B.~Lakshminarayanan, A.~Pritzel, and C.~Blundell, ``{S}imple and scalable
  predictive uncertainty estimation using deep ensembles,'' in \emph{NIPS},
  2017.

\bibitem{akai2020hybrid}
N.~Akai, T.~Hirayama, and H.~Murase, ``{H}ybrid localization using model- and
  learning-based methods: {Fusion of Monte Carlo and E2E localizations via
  importance sampling},'' in \emph{ICRA}, 2020.

\bibitem{gao2020multi}
P.~Gao, R.~Guo, H.~Lu, and H.~Zhang, ``{M}ulti-view sensor fusion by
  integrating model-based estimation and graph learning for collaborative
  object localization,'' \emph{ICRA}, 2021.

\bibitem{barfoot2014batch}
T.~D. Barfoot, C.~H. Tong, and S.~S{\"a}rkk{\"a}, ``Batch continuous-time
  trajectory estimation as exactly sparse gaussian process regression.'' in
  \emph{Robotics: Science and Systems}, vol.~10, 2014.

\bibitem{greff2016lstm}
K.~Greff, R.~K. Srivastava, J.~Koutn{\'\i}k, B.~R. Steunebrink, and
  J.~Schmidhuber, ``{LSTM}: A search space odyssey,'' \emph{IEEE T NEURAL
  NETWOR}, vol.~28, no.~10, pp. 2222--2232, 2016.

\bibitem{velivckovic2017graph}
P.~Veli{\v{c}}kovi{\'c}, G.~Cucurull, A.~Casanova, A.~Romero, P.~Lio, and
  Y.~Bengio, ``Graph attention networks,'' \emph{International Conference on
  Representation Learning}, 2018.

\bibitem{ramachandran2017searching}
P.~Ramachandran, B.~Zoph, and Q.~V. Le, ``Searching for activation functions,''
  \emph{ICLR Workshop}, 2018.

\bibitem{lakshminarayanan2016simple}
B.~Lakshminarayanan, A.~Pritzel, and C.~Blundell, ``Simple and scalable
  predictive uncertainty estimation using deep ensembles,'' \emph{NIPS}, 2017.

\bibitem{li2015kalman}
Q.~Li, R.~Li, K.~Ji, and W.~Dai, ``Kalman filter and its application,'' in
  \emph{IINS}, 2015, pp. 74--77.

\bibitem{zhang2020single}
Q.~Zhang and A.~B. Chan, ``{S}ingle-frame based deep view synchronization for
  unsynchronized multi-camera surveillance,'' \emph{ArXiv}, 2020.

\bibitem{zhang2004extrinsic}
Q.~Zhang and R.~Pless, ``Extrinsic calibration of a camera and laser range
  finder,'' in \emph{IROS}, 2004.

\bibitem{brahmbhatt2018geometry}
S.~Brahmbhatt, J.~Gu, K.~Kim, J.~Hays, and J.~Kautz, ``{G}eometry-aware
  learning of maps for camera localization,'' in \emph{CVPR}, 2018.

\bibitem{kendall2015posenet}
A.~Kendall, M.~Grimes, and R.~Cipolla, ``{PoseNet: A} convolutional network for
  real-time {6-DOF} camera relocalization,'' in \emph{ICCV}, 2015.

\bibitem{gaoregularized}
P.~Gao, R.~Guo, H.~Lu, and H.~Zhang, ``{R}egularized graph matching for
  correspondence identification under uncertainty in collaborative
  perception,'' in \emph{RSS}, 2020.

\bibitem{khoshelham2011accuracy}
K.~Khoshelham, ``{A}ccuracy analysis of {K}inect depth data,'' in \emph{ISPRS
  workshop laser scanning}, 2011.

\bibitem{boyd2011distributed}
S.~Boyd, N.~Parikh, and E.~Chu, \emph{Distributed optimization and statistical
  learning via the alternating direction method of multipliers}.\hskip 1em plus
  0.5em minus 0.4em\relax Now Publishers Inc, 2011.

\bibitem{ji2017surfacenet}
M.~Ji, J.~Gall, H.~Zheng, Y.~Liu, and L.~Fang, ``{SurfaceNet: An} end-to-end
  {3D} neural network for multiview stereopsis,'' in \emph{ICCV}, 2017.

\bibitem{godard2017unsupervised}
C.~Godard, O.~Mac~Aodha, and G.~J. Brostow, ``{Unsupervised} monocular depth
  estimation with left-right consistency,'' in \emph{CVPR}, 2017.

\end{thebibliography}
\end{document}